\documentclass[conference]{IEEEtran}
\usepackage{times}

\usepackage{xcolor}
\usepackage{tcolorbox}
\usepackage{color,soul}
\usepackage[utf8]{inputenc}
\usepackage{amsmath}
\usepackage{amsfonts} 
\usepackage{booktabs}
\usepackage{float}
\usepackage{bbm}
\usepackage{xspace}
\usepackage{graphics}
\usepackage{subcaption}
\usepackage{graphicx}
\usepackage{multirow}
\usepackage{multicol}
\usepackage{url}
\usepackage{hyperref}

\newcommand{\ignore}[1]{}

\newcommand{\ba}{\begin{array}}
\newcommand{\ea}{\end{array}}
\newcommand{\bc}{\begin{center}}
\newcommand{\ec}{\end{center}}
\newcommand{\be}{\begin{enumerate}}
\newcommand{\ee}{\end{enumerate}}
\newcommand{\bea}{\begin{eqnarray}}
\newcommand{\eea}{\end{eqnarray}}
\newcommand{\beas}{\begin{eqnarray*}}
\newcommand{\eeas}{\end{eqnarray*}}
\newcommand{\beq}{\begin{equation}}
\newcommand{\eeq}{\end{equation}}
\newcommand{\bfig}{\begin{figure}}
\newcommand{\efig}{\end{figure}}
\newcommand{\bi}{\begin{itemize}}
\newcommand{\ei}{\end{itemize}}
\newcommand{\bpic}{\begin{picture}}
\newcommand{\epic}{\end{picture}}
\newcommand{\btabular}{\begin{tabular}}
\newcommand{\etabular}{\end{tabular}}
\newcommand{\btable}{\begin{table}}
\newcommand{\etable}{\end{table}}

\newcommand{\es}{\vfill
                 \rule[-6mm]{170mm}{0.7mm} \\
                 \redw{{\tiny
		  \hfill S-\theslide}}
                 \end{slide}}

\newcommand{\matxx}[1]{{\mathtt #1}}
\newcommand{\vecXX}[1]{{\mathbf {#1}}}

\newcommand{\argmin}{\operatornamewithlimits{arg\ min}}

\def \hbar {{\bar{h}}}

\def \vecg {{\vecXX{g}}}

\def \vecn {{\vecXX{n}}}

\def \vecp {{\vecXX{p}}}

\def \vecr {{\vecXX{r}}}

\def \vecx {{\vecXX{x}}}
\def \vecy {{\vecXX{y}}}

\def \matA {{\matxx{A}}}



\usepackage[numbers]{natbib}
\usepackage{multicol}

\begin{document}

\title{
iSDF: Real-Time Neural Signed Distance Fields\\for Robot Perception
}

\author{
Joseph Ortiz$^{1,2}$ \hspace{3mm}
Alexander Clegg$^2$ \hspace{3mm}
Jing Dong$^3$ \hspace{3mm}
Edgar Sucar$^1$ \\
David Novotny$^2$ \hspace{3mm}
Michael Zollhoefer$^3$ \hspace{3mm}
Mustafa Mukadam$^2$\\[5pt]
$^1$Imperial College London \hspace{3mm}
$^2$Meta AI \hspace{3mm}
$^3$Reality Labs Research
\vspace{-5mm}
}

\twocolumn[{
	\renewcommand\twocolumn[1][]{#1}
	\maketitle
	\thispagestyle{empty}
	\begin{center}
		\vspace{-5mm}
		\centering
		\includegraphics[width=\linewidth]{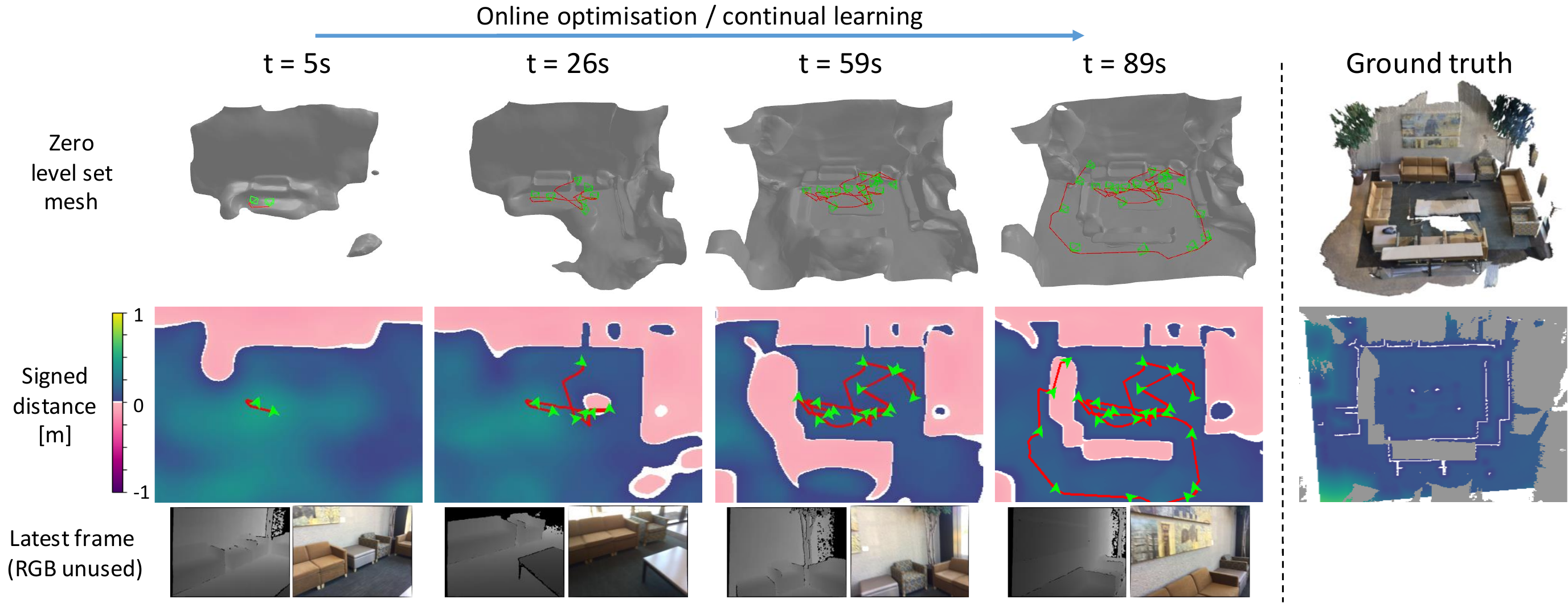}
		\captionof{figure}{We present iSDF, a system for real-time signed distance field reconstruction that optimises a randomly initialised network to regress to the signed distance for input 3D coordinates based on posed depth images from a live camera stream. To train the network in real-time we develop a batch-based self-supervision method.
		This network can be queried online to obtain collision costs and gradients for use by downstream planners in domains from navigation to manipulation. The trajectory (red) of the camera (green) is visualised on top of the zero level set (top row) and SDF slice (middle row) generated by iSDF over time (left to right).}
		\vspace{-1mm}
		\label{fig:teaser}
	\end{center}
}]

\begin{abstract}
%
%
We present iSDF, a continual learning system for real-time signed distance field (SDF) reconstruction. Given a stream of posed depth images from a moving camera, it trains a randomly initialised neural network to map input 3D coordinate to approximate signed distance.
The model is self-supervised by minimising a loss that bounds the predicted signed distance using the distance to the closest sampled point in a batch of query points that are actively sampled.
In contrast to prior work based on voxel grids, our neural method is able to provide adaptive levels of detail with plausible filling in of partially observed regions and denoising of observations, all while having a more compact representation.
In evaluations against alternative methods on real and synthetic datasets of indoor environments, we find that iSDF produces more accurate reconstructions, and better approximations of collision costs and gradients useful for downstream planners in domains from navigation to manipulation.
Code and video results can be found at our project page: \color{blue}{\url{https://joeaortiz.github.io/iSDF/}}.
\end{abstract}

\IEEEpeerreviewmaketitle

\vspace{-1mm}
\section{Introduction}


Robot perception is often the forgotten middle child.
One instantiation of this syndrome occurs with signed distance fields (SDFs), a map representation common in both robotics and vision.
SDFs are scalar fields that associate each point in space with the signed distance to the closest surface point, therefore encoding the surface as the zero level set.
In robotics, SDFs are a common environment representation used for collision avoidance in motion planning \cite{Zucker:etal:IJRR:2013}, however they are usually assumed as given a priori or too expensive to compute in real-time.
On the other hand, in vision, truncated signed distance fields (TSDF) are common in SLAM systems \cite{Newcombe:etal:ISMAR2011, Newcombe:etal:CVPR2015}, but there is little work on reconstructing non-truncated SDFs in room scale environments.
In order to close this gap, in this work, we focus on the problem of real-time SDF reconstruction.



Within robot motion planning, signed distance fields are the prevailing map representation used in trajectory optimisation \cite{Zucker:etal:IJRR:2013, Toussaint:ICML:2009, Schulman:etal:IJR:2014, Dong:etal:RSS:2016, Mukadam:etal:IJR:2018} where a trajectory is found by minimising an objective balancing smoothness and collision avoidance costs.
Collision costs are dependent on the environment and robot state and in practise SDFs are a common intermediate representation used to mitigate challenges in building generic cost fields for high dimensional robots.
Surface representations, such as occupancy maps or TSDFs, are common in SLAM systems but impractical for planning since trajectory optimisation requires many cost queries. It is more efficient to precompute a cheap-to-evaluate cost field, like a SDF, than to compute costs on the fly from a surface reconstruction.
Although truncated SDFs can be rapidly constructed by fusing depth measurements, producing the full SDF is far more challenging as fusion alone leads to large errors away from surfaces.

Prior work on real-time reconstruction of non-truncated SDFs operates in two stages, first reconstructing surface geometry and then transforming it to the SDF using wavefront propagation algorithms \cite{oleynikova:etal:IROS2017, Han:etal:IROS2019}.
These methods are based on voxel grids and are limited to a resolution of 5cm due to the cost of wavefront propagation. This is the case for both efficient CPU-based propagation algorithms \cite{oleynikova:etal:IROS2017, Han:etal:IROS2019} as well as for GPU-based brute force propagation algorithms \cite{Felzenszwalb:Huttenlocher:2004}.

Neural fields provide an alternative paradigm to voxel grids for modelling scenes \cite{Park:etal:CVPR2019, mescheder2019occupancy, Mildenhall:etal:ECCV2020}.
Based on a multi-layer perceptron (MLP) that maps a 3D coordinate to occupancy, these models can be optimised from scratch to accurately fit a specific scene without prior training.
Recent work has shown that neural fields can reconstruct highly accurate 3D geometry and that they can be trained in real-time as part of a SLAM system \cite{Sucar:etal:ICCV2021, Zhu:etal:ARXIV2021}.

Building on these advances, we present iSDF (\textit{incremental Signed Distance Fields}), a system for real-time SDF reconstruction that is based on a neural network that regresses input 3D coordinates to signed distance.
Unlike iMAP \cite{Sucar:etal:ICCV2021}, which reconstructs a neural radiance field as part of a SLAM system, we focus on mapping with a neural signed distance field and develop novel self-supervision and sampling strategies.

More specifically, given a stream of posed depth images, a randomly initialised network is trained in a continual learning fashion via actively replaying past observations.
Our approach is simple, yet effective, and runs in real-time (33ms per step); we employ a single small network, sparse sampling, and do not require pretraining.
During online operation, the model is queried at points sampled along back-projected rays in the frames and the network weights are optimised to minimise a loss on both the predicted signed distance and its spatial gradient.
Our loss bounds the SDF prediction with the distance to the closest surface point in the batch and enables the learning of signed distances far from surfaces without the need for wavefront propagation.

iSDF inherits the positive characteristics of neural fields. In contrast to prior work based on voxel grids, iSDF is:
(i) efficient - it can seamlessly allocate memory capacity to model different parts of an environment with different levels of detail, and
(ii) predictive - it can denoise and consolidate noisy measurements and sensibly fill in gaps in partially observed regions; all while having a more compact representation.

We show that iSDF outperforms prior work in terms of SDF accuracy on both synthetic and real datasets in indoor environments.
iSDF achieves an SDF error of less than 6cm for all our sequences and is the only method that can reconstruct watertight meshes of the zero level set.
We also evaluate on computing collision costs and gradients from signed distance and find that iSDF outperforms alternatives on these metrics, demonstrating its utility for downstream planners in domains from navigation to manipulation.




\vspace{-1mm}
\section{Related Work}

Since the seminal works \cite{Park:etal:CVPR2019, mescheder2019occupancy, Mildenhall:etal:ECCV2020}, neural fields have become a prevailing representation for modelling both 3D objects and full scenes.
In particular DeepSDF \cite{Park:etal:CVPR2019} inspired a large number of works on reconstructing object surfaces as the zero level set of neural signed distance fields.
Along this direction, \citet{Gropp:etal:ICML2020} introduce Eikonal regularisation to supervise the SDF learning for points in free space while \citet{Zhang:etal:ARXIV2021} use the surface normals to supervise free space points.
A number of works address differentiable rendering to learn SDFs from images \cite{Jiang:etal:CVPR2020, Yariv:etal:NEURIPS2021, Peng:etal:ARXIV2021, Liu:etal:CVPR2022} while Atzmon et al.~\cite{Atzmon:Lipman:CVPR2020, Atzmon:Lipman:ICLR:2021} focus on learning SDFs from raw pointcloud data.

Truncated SDFs have commonly been used in traditional SLAM systems \cite{Newcombe:etal:ISMAR2011, Newcombe:etal:CVPR2015}, however more recently a number of works have used neural fields to represent the TSDFs of room-scale environments \cite{azinovic:etal:ARXIV2021, Yan:etal:ICCV2021, Sitzmann:etal:NIPS2020}.
To the best of our knowledge there is no prior work tackling non-truncated SDF reconstruction in room scale environments.

Neural fields showed great early promise with offline reconstruction tasks. Later, iMAP \cite{Sucar:etal:ICCV2021} was the first work to show that these models can be trained in a real-time continual learning setting as part of a SLAM system.
Our work is strongly inspired by iMAP and indeed the active sampling and keyframe selection in iSDF are based on iMAP.
NICE-SLAM \cite{Zhu:etal:ARXIV2021} builds on iMAP by proposing to use a voxel grid of neural fields instead of a single global model.
Although not a real-time system, \citet{Yan:etal:ICCV2021} investigates offline continual learning for neural fields.

The most closely related prior works that tackle real-time SDF reconstruction from posed depth images are Voxblox \cite{oleynikova:etal:IROS2017} and FIESTA \cite{Han:etal:IROS2019}.
These methods operate in two phases and output a voxel grid of SDF values.
First, depth images are fused into a surface representation (TSDF or occupancy grid) before the full SDF is computed via efficient wavefront propagation algorithms.
The key to the efficiency is to only emanate wavefronts from updated voxels with the propagation being done by a breadth first search (BFS).
We find that Voxblox and FIESTA perform comparably and in evaluations show that iSDF produces more accurate and complete SDFs than Voxblox.

Other notable works tackle SDF reconstruction in dynamic environments with composite fields \cite{Finean:etal:ICAPS2021} and in very large environments using submaps \cite{Reijgwart:etal:RAL2019}.
Lastly, other related directions tackle learning cost fields for motion planning \cite{Chaplot:etal:ICML2021, Das:Yip:TRO2020, Verghese:eta:RAL2022} from SDFs \cite{Driess:etal:CoRL2022, Hayne:etal:ICRA2016}, navigating in a neural radiance fields \cite{Adamkiewicz:etal:ARXIV2021} and learning neural fields for articulated \cite{Jiang:etal:ARXIV2022} or deformable objects \cite{Wi:etal:ARXIV2022} for manipulation. 

\section{Signed distance fields}

A signed distance field is a scalar field which associates every point in space with the signed distance to the closest surface.
The distance is assigned a negative sign for points inside the surface and takes a positive sign outside of the surface.
When representing three dimensional geometry, the signed distance function maps a 3D coordinate $\vecx$ to the scalar signed distance value $f : \mathbb{R}^3 \rightarrow \mathbb{R}$.
The surface $\mathcal{S}$ is the zero level-set of the field:
\begin{equation}
    \mathcal{S} = \{ \vecx \in \mathbb{R}^3 ~ \vert ~ f(\vecx) = 0 \}
    ~.
\end{equation}

A signed distance field has several notable properties that we will use to construct our loss function (Sec.~\ref{subsec:loss}):
(i) $f$ is differentiable almost everywhere. It is not differentiable at points where there are multiple equidistant closest surface points,
(ii) where the gradient exists, $-\nabla_\vecx f$ points towards the closest surface and on the surface the gradient is equal to the surface normal: $\nabla_\vecx f = \hat{\vecn}$, and
(iii) the gradient vector satisfies the Eikonal equation: $\rvert \nabla_\vecx f \lvert = 1$, as moving a distance $\delta x$ along the gradient vector increases the distance to the closest surface by $\delta x$.

\section{iSDF - Real-time SDF reconstruction}
\begin{figure}[t]
    \centering
    \includegraphics[width=\linewidth]{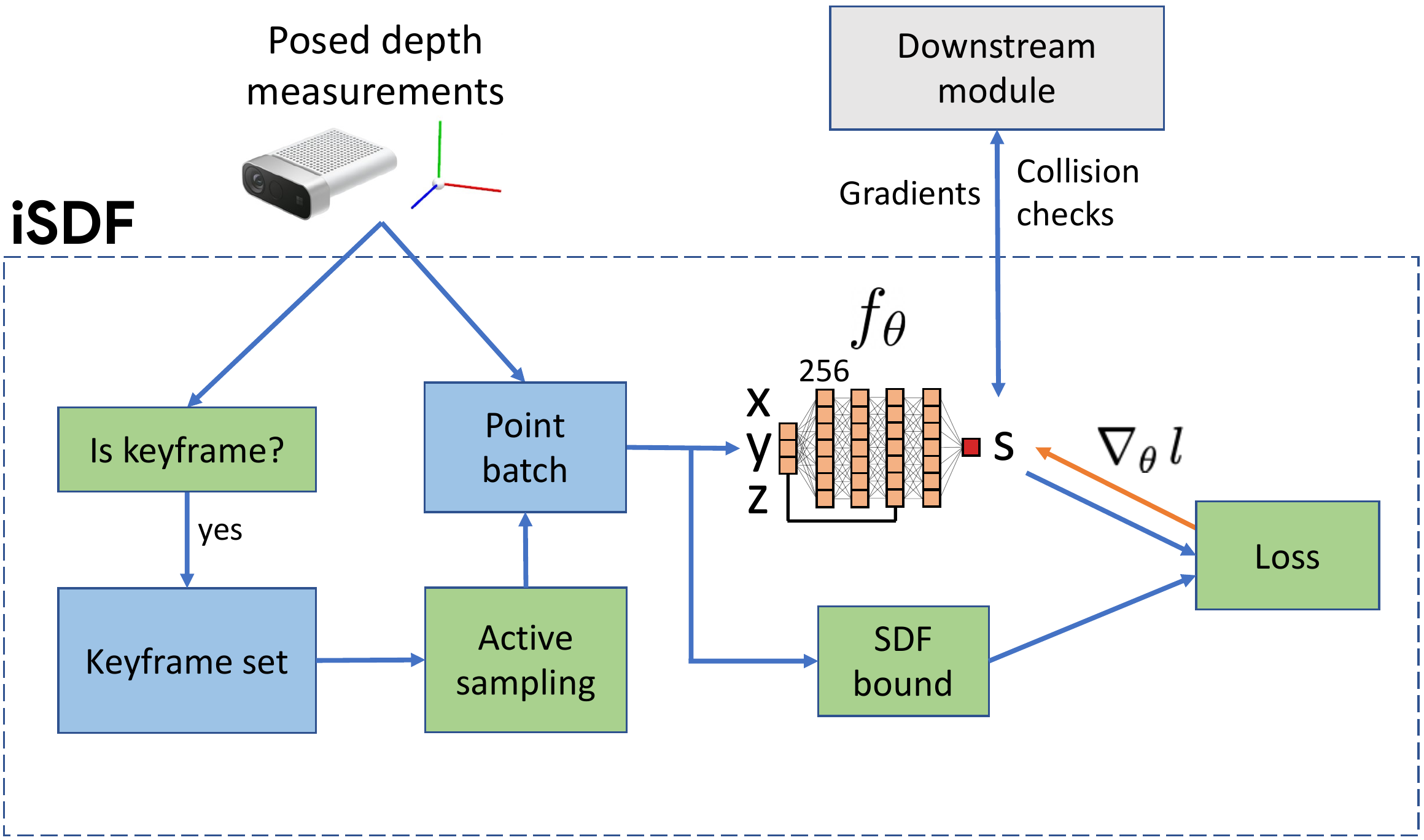}
    \caption{\textbf{System diagram.} iSDF's MLP regresses input 3D coordinates $\vecx = (x, y, z)$ to the signed distance $s = f_\theta(\vecx)$.}
    \label{fig:system_diagram}
    \vspace{-5mm}
\end{figure}
We present iSDF a system for real-time SDF reconstruction, that takes as input a stream of posed depth images captured by a moving camera and, during online operation, learns a function that approximates the true signed distance field of the environment.
An overview of our system is given in Fig.~\ref{fig:system_diagram}.
We focus on mapping rather than the full SLAM problem and therefore assume there is an external tracking module.
The signed distance function is modelled by a multilayer perceptron (MLP), that maps a 3D coordinate $\vecx$ to the signed distance value $s$ at that point:
$f(\vecx ; \theta) = s$ (Sec.~\ref{subsec:architecture}).
The model is initialised with random weights and is optimised in real-time with respect to incoming measurements.

At each iteration, we select a subset of frames, i.e. a few of the posed depth images.
Given the continual learning setting, the frames are selected via active sampling from a sparse set of representative keyframes to mitigate catastrophic forgetting.
For each frame, we sample both random pixels and depths along the corresponding back-projected rays to obtain a batch of points (Sec. \ref{subsec:sampling}).
The SDF model is queried at the sample points and we compute a loss that is minimised by refining the network weights with back-propagation.
The loss supervises both the predicted signed distance and its spatial gradient (Sec.~\ref{subsec:loss}) and is formed by bounding the predicted signed distance with the distance to the closest surface point in the batch (Sec.~\ref{subsec:closest_surf_point}).

\subsection{Network architecture}
\label{subsec:architecture}
We model the signed distance function using an MLP with 4 hidden layers, 256 activations each, and softplus activations in all intermediate layers.
As in NeRF \cite{Mildenhall:etal:ECCV2020}, we apply positional embedding to the 3D coordinates before feeding them to the network.
This embedding transforms the input coordinates to a high dimensional vector using periodic activation functions and is crucial for reconstructing high frequency signals.
We employ the ``off-axis" positional embedding  $\gamma(\vecx) = [\sin(2^0 \matA \vecx), \cos(2^0 \matA \vecx), \dots \,, \sin(2^L \matA \vecx), \cos(2^L \matA \vecx)]$ with $L=5$ from \citet{Barron:etal:arxiv:2020}, where the rows of $\matA \in \mathbb{R}^{21 \times 3}$ are the outward facing unit-norm vertices of a twice-tessellated Icosahedron.
The positional embedding is also concatenated to the third layer of the network. 

\subsection{Active sampling}
\label{subsec:sampling}
The sampling procedure, broadly follows iMAP \cite{Sucar:etal:ICCV2021} on the level of frames and NeRF \cite{Mildenhall:etal:ECCV2020} on the level of points.
Optimising the network with respect to all incoming frames would be both computationally infeasible and redundant.
Therefore, as in iMAP \cite{Sucar:etal:ICCV2021}, we  maintain a sparse set of keyframes which are replayed via active sampling to avoid catastrophic forgetting.
The first frame is always added to the keyframe set and subsequent frames are added based on the information gain metric from iMAP \cite{Sucar:etal:ICCV2021} that is computed using a frozen copy of the network when the last keyframe was added.
At each iteration, we select 5 frames - the 2 latest frames and 3 keyframes from the keyframe set sampled from a distribution that is formed by normalising the running total losses of each keyframe.
New frames are sampled for 10 iterations before checking for a more recent frame.

Given each selected frame, which comprises of a camera pose $T_{WC}$ and a depth image $D$, we sample query points within the viewing frustum.
First, we randomly sample 200 pixel coordinates $[u, v]$ in the depth image.
Then, the viewing direction is computed using the camera intrinsic matrix $K$ and transformed into world coordinates: $\vecr = T_{WC} K^{-1} [u, v]$.

For each ray, we sample $N + M + 1$ depths values $d_i$ along the ray.
The depth values are made up of $N$ stratified samples in the range $[d_{min}, D[u, v] + \delta]$ and $M$ samples from the Gaussian distribution $\mathcal{N}(D[u, v], \sigma_s^2)$ to provide more supervision around the surface where the SDF is harder to model.
Additionally, we always include the sample at the surface given by the depth map by including $d_i = D[u, v]$.
These depth samples, along with the viewing directions, specify the sample points along the rays: $\vecx_i = d_i \vecr$.
The batch of points is the set of sample points from all selected frames.

In all experiments, we set $N=20$, $M=8$, $d_{min} = 7cm$, $\delta = 10cm$ and $\sigma_s = 10cm$.

\subsection{Approximating the closest surface point}
\label{subsec:closest_surf_point}
At each iteration, given the batch of points, we query the network to give the predicted signed distances:
$\hat{s} = f(\vecx; \theta)$.
Ideally, we would like to supervise these predictions using the true signed distance values:
\begin{equation}
    s (\vecx) = \text{sgn}(D[u,v] - d) \,\inf_{\vecy \in \mathcal{S}} \lvert \vecx - \vecy \rvert
    ~,
\end{equation}
where $\inf$ denotes the infimum, $\text{sgn}$ the sign function and $\mathcal{S}$ is the set of all true surface points.
It would however be computationally infeasible to compute the distance to all surface points and would only yield an approximate signed distance due to the geometry being partially observed. 

Instead of computing the distance to all surface points, we propose to instead use the distance to a single carefully chosen nearby surface point $\vecp$.
The nearby surface point is unlikely to be the true closest surface point so will yield a signed distance value with larger absolute value than the true signed distance.
In this way, computing the distance to a nearby surface point provides a bound $b (\vecx, \vecp)$ on the signed distance:
\begin{equation}
    b (\vecx, \vecp) = \text{sgn}(D[u,v] - d) \, \lvert \vecx - \vecp \rvert
    ~,
\end{equation}
where $\lvert s(\vecx) \rvert \leq \lvert b(\vecx, \vecp) \rvert$, and with equality when $\vecp$ is the closest surface point to $\vecx$.

The closer the nearby surface point is to the query point, the tighter the bound $b (\vecx, \vecp)$ is on the true signed distance. We explore a number of methods for choosing the nearby surface point $\vecp$ that trade of increasingly tighter bounds for additional computation.

The cheapest option we explore is to use the ray intersection with the surface: $\vecp = D[u,v] \, \vecr$.
In this case, the bound is simply the signed distance along the ray between the query point depth and the measured depth:
$b (\vecx, D[u,v] \, \vecr) = \text{sgn}(D[u,v] - d) \lvert D[u, v] - d \rvert$.
If the surface is perpendicular to the ray / viewing direction $\vecr$, then this bound is more likely to be tight while if these vectors are not perpendicular we can always find a new surface point that is closer to the query point by moving a small distance along the surface.

\begin{figure}
    \centering
    \includegraphics[width=\linewidth]{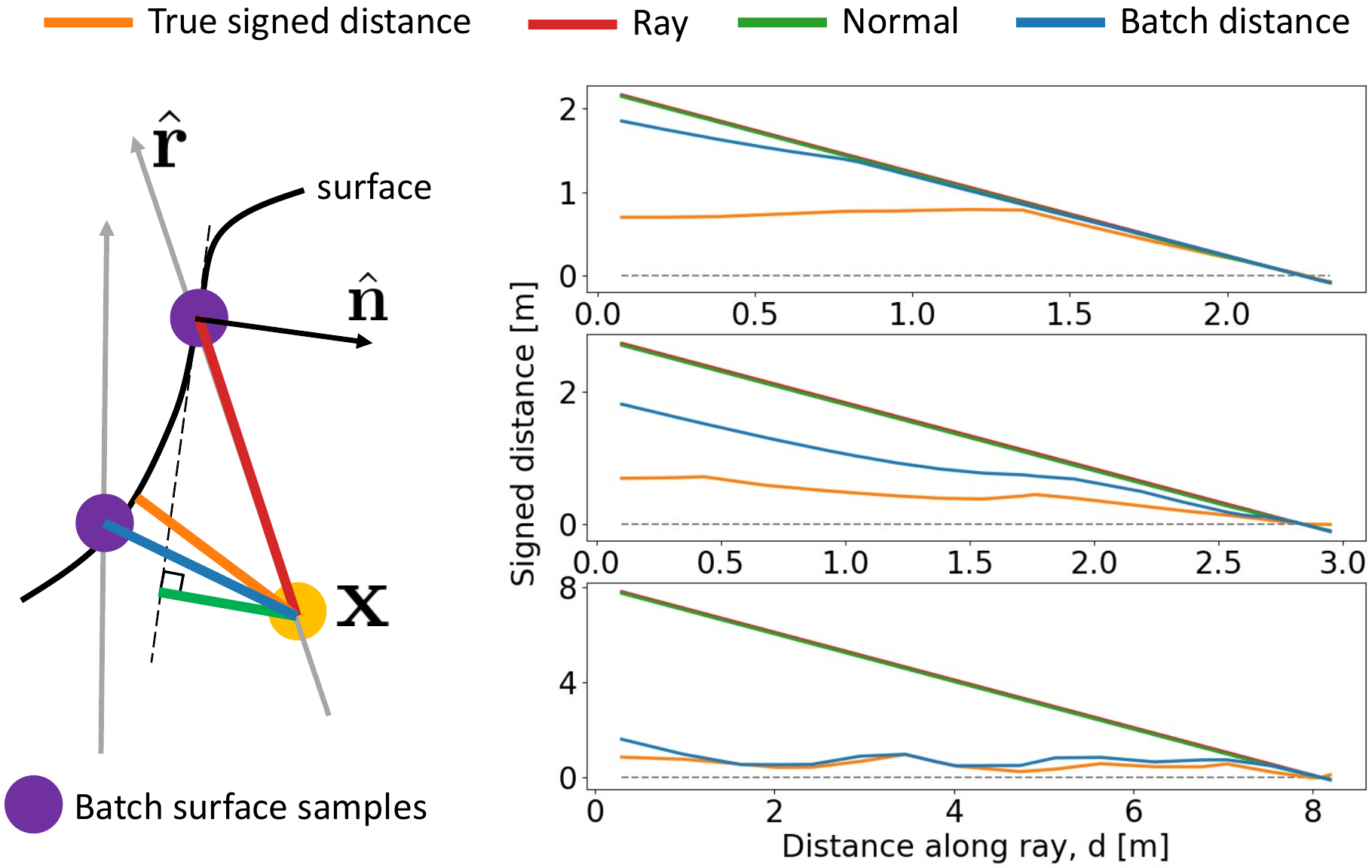}
    \caption{\textit{Left:} Different methods for computing the bound on the SDF prediction for sample point $\vecx$ (orange). \textit{Right:} The bound for each method for sample points along a ray. Each plot shows a different ray and the distance along the ray is measured from the camera centre. 
    The batch distance provides the tightest bounds on the true signed distance.}
    \label{fig:target_sdf_diagram}
    \vspace{-1mm}
\end{figure}

To approximately account for this, we can apply a correction to the bound using the angle between the surface normal and the viewing direction. As illustrated in Fig.~\ref{fig:target_sdf_diagram}, if we assume the surface is planar around the ray surface intersection, then the normal adjusted bound is:
$b (\vecx, \hat{\vecr}, \hat{\vecn}) = -\hat{\vecr} \cdot \hat{\vecn} \; (d - D[u, v])$, where $\hat{\vecr} = \frac{\vecr}{\rvert \vecr \lvert}$.
As we assume that the surface is only locally planar, we apply the normal correction to points less than 30cm from the ray surface intersection, and use the ray bound elsewhere. To apply this correction, we must first compute the surface normal as the spatial gradient of the depth image.

The last method we consider is brute-force computation of the distance to the closest surface point in the batch.
This gives the following bound:
\begin{equation}
    b(\vecx, \mathcal{P}) = \text{sgn}(D[u,v] - d) \, \min_{\vecp \in \mathcal{P}} \rvert \vecx - \vecp \lvert
    ~,
\end{equation}
where $\mathcal{P}$ is the set of surface points in the batch that are sampled at the measured depth along the ray (i.e. $\vecp = D[u,v] \vecr$).

For all methods, we can also compute an approximate SDF gradient at $\vecx$ using the fact that $- \nabla_\vecx f (\vecx)$ points towards the closest surface.
For example, using the closest batch surface point, the approximate unnormalised gradient is:
\begin{equation}
    \vecg (\vecx, \mathcal{P}) = \text{sgn}(D[u, v] - d) \, . \, \big( \vecx - \argmin_{\vecp \in \mathcal{P}} \rvert \vecx - \vecp \lvert \big)
    ~.
\end{equation}

The three methods we consider above for computing the bound are illustrated on the left of Fig.~\ref{fig:target_sdf_diagram}.
On the right, we plot the bound for each method at query points along different rays.
The ray distance clearly provides very weak bounds and normal correction leads to minimal improvements.
On the other hand, computing the batch distance can provide tight bounds depending on whether the batch has good surface coverage.
In practice, we find that the batch distance method yields the best performance at little extra computation cost (see Sec.~\ref{subsec:bounds_ablation}) and thus employ this method in all experiments.

\begin{figure}
    \centering
    \includegraphics[width=\linewidth]{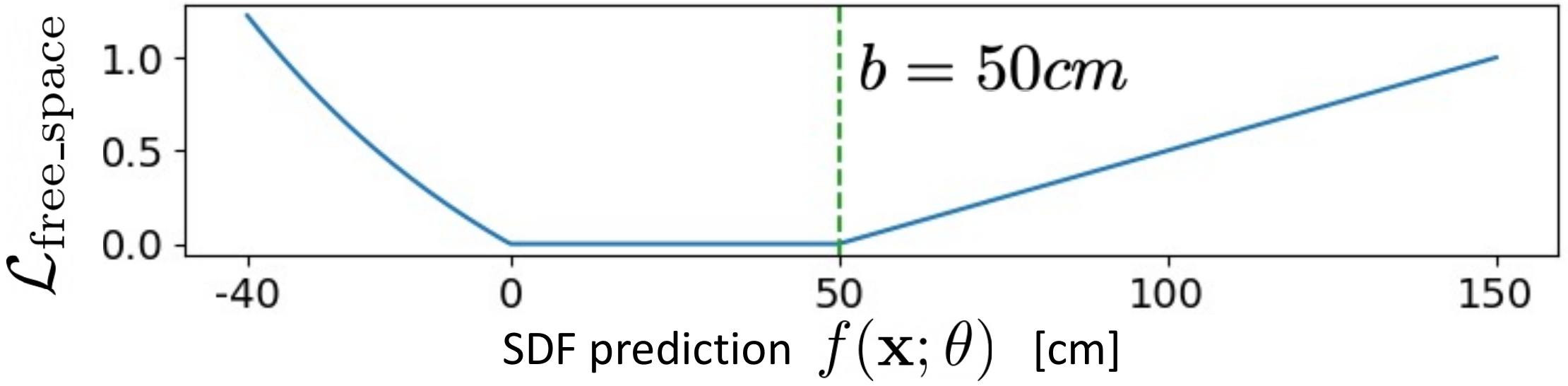}
    \caption{\textbf{Free space loss.} $\mathcal{L_{\text{free\_space}}} ( f(\vecx; \theta), b)$ for target SDF.}
    \label{fig:free_space_loss}
    \vspace{-2mm}
\end{figure}

\subsection{Loss for self-supervised continual learning}
\label{subsec:loss}

\textbf{SDF loss.}
Using the bound on the predicted SDF value, we construct a loss $\mathcal{L_{\text{free\_space}}}$ in Fig.~\ref{fig:free_space_loss} for points in free space.
The loss is zero if the prediction is positive and less than $b$, linear in the prediction when it exceeds the upper bound $b$, and exponential for negative predictions (we set $\beta = 5$):
\begin{equation}
    \mathcal{L_{\text{free\_space}}} ( f(\vecx; \theta), b) = \max \big(0, ~ e^{-\beta f(\vecx_i; \theta)} - 1, ~f(\vecx_i; \theta) - b \big) 
    ~.
\end{equation}
Close to the surface, we expect the bound to be very tight and so instead of applying the free space loss, we directly supervise the SDF prediction to take the bound value via an $L1$ loss:
\begin{equation}
    \mathcal{L_{\text{near\_surf}}} ( f(\vecx; \theta), b) = \vert f(\vecx_i; \theta) - b \vert
    ~.
\end{equation}
Applying the near surface loss within a truncation region $t$ and the free space loss elsewhere, our full SDF loss is:
\begin{equation}
    \mathcal{L_{\text{sdf}}} ( f(\vecx; \theta), b) = 
    \begin{cases}
        \lambda_\text{surf} \mathcal{L_{\text{near\_surf}}} &\mbox{if } \vert D[u, v] - d \vert  \leq t \\
        \mathcal{L_{\text{free\_space}}} &\mbox{otherwise.}
    \end{cases}
\end{equation}

\textbf{Gradient loss.}
The gradient of the SDF prediction can be efficiently computed via automatic differentiation.
We therefore also supervise the SDF gradient prediction using our approximated gradient $\vecg$, computed using the closest surface point in the batch.
We introduce a loss penalising the cosine distance between the two vectors:
\begin{equation}
    \mathcal{L_{\text{grad}}}( \nabla_{\vecx} f(\vecx; \theta), \vecg) = 
    1 - \frac{\nabla_{\vecx} f(\vecx; \theta) \cdot \vecg}{\| \nabla_{\vecx} f(\vecx; \theta) \|  \| \vecg \|}
    ~.
\end{equation}
For surface point samples, we replace $\vecg$ with the surface normal computed from the depth image.

\textbf{Eikonal regularisation.}
Without very dense supervision, the above losses alone do not result in the model learning a valid signed distance field which satisfies the Eikonal equation almost everywhere.
Consequently, following \citet{Gropp:etal:ICML2020} we apply Eikonal regularisation via the loss:
\begin{equation}
    \mathcal{L_{\text{eik}}} ( f(\vecx; \theta)) = 
    \begin{cases}
         \lvert\, \|\nabla_{\vecx} f(\vecx; \theta)\| - 1 \rvert &\mbox{if } \vert D[u, v] - d \vert  \geq a \\
        0 &\mbox{otherwise.}
    \end{cases}
\end{equation}
Different to prior work, we only regularise points a distance greater than $a=10cm$ from the ray surface intersection.
We find this improves performance, as gradient discontinuities that do not satisfy the Eikonal equation are far more common close to surfaces where nearby points often have different closest surfaces.
The regularisation has the effect of propagating the field out from near the surfaces (where there is the strongest signal from the SDF loss) into free space.
In this sense, it can be thought of as playing a similar role to the wavefront propagation algorithms. 

The network parameters are optimised to minimise the loss:
\begin{equation}
    l(\theta) = \mathcal{L_{\text{sdf}}} + \lambda_\text{grad} \, \mathcal{L_{\text{grad}}} + \lambda_\text{eik} \, \mathcal{L_{\text{eik}}}
    ~.
\end{equation}
We set $\lambda_\text{surf} = 5$, $\lambda_\text{grad} = 0.02$ and $\lambda_\text{eik} = 0.25$ and the weights are optimised using the Adam optimiser with learning rate $0.0013$ and weight decay $0.012$.

\section{Experimental Setup}

\subsection{Datasets}

\textbf{ReplicaCAD \cite{Szot:etal:ARXIV2021}.}
We experiment with 6 synthetic sequences from the ReplicaCAD dataset \cite{Szot:etal:ARXIV2021} generated by simulating the measurements recorded by a camera mounted on a mobile manipulator.
For two different room configurations, we generate 3 sequences which each aim to evaluate different properties of the reconstruction.
For each room we generate 1) a navigation sequence (\textit{nav}) in which the robot explores the majority of the room by navigating between waypoints, 2) an object reconstruction sequence (\textit{obj}) in which the robot approaches and looks directly at two large objects (e.g. a lamp or bag), and 3) a manipulation sequence (\textit{mnp}) in which the robot navigates close to a small object (e.g. a bowl) with the intent to reach out and grasp it.
We apply the noise model from \citet{Choi:etal:CVPR2015} to all depth images to account for disparity-based quantisation effects, realistic high-frequency noise, and low-frequency distortions.

\textbf{ScanNet \cite{Dai:etal:CVPR2017}.}
We use 3 longer and 3 shorter randomly chosen sequences from the ScanNet dataset.
These sequences were captured using a handheld RGB-D camera.

\textbf{Ground truth SDF computation.}
For each sequence, we pre-compute a voxel grid of ground truth SDF values with a voxel resolution of 1cm.
To evaluate the real-time SDF reconstruction, this grid is interpolated to produce values at query points.
The SDF is computed by first voxelizing the mesh into an occupancy grid, then computing the euclidean distance transform of the occupancy and inverse occupancy grid, before subtracting the latter from the former to give the SDF.
For the ReplicaCAD sequences, we compute separately the SDF for an empty room and for each object in the room.
These are then composed using the min-operation to give the full room SDF.
SDF computation is more challenging for the ScanNet sequences as the mesh is not watertight, making it difficult to determine whether a point is in free space or contained within a surface.
Consequently, for the ScanNet sequences, we only evaluate in visible regions, where the SDF attains strictly positive values.

\subsection{Comparisons}

We compare iSDF against two voxel-based methods for real-time SDF reconstruction that employ a two stage pipeline of first reconstructing the surface via fusing depth images and then transforming to SDF. For both methods the second stage is the dominant computational cost.

\textbf{Voxblox \cite{oleynikova:etal:IROS2017}.}
Voxblox is a CPU-only algorithm that first runs a fast TSDF fusion before efficiently computing the SDF by propagating wavefronts only from updated voxels.
The voxel size is set to 5.5cm such that SDF updates take around half a second for room scale environments; any higher resolution would make Voxblox too slow for real-time planning.
Voxblox employs complex data structures for efficient wavefront propagation and would therefore be challenging to parallelise on a GPU.

\textbf{KinectFusion+.}
To the best of our knowledge there are no methods that leverage GPUs for real-time SDF reconstruction. Consequently, we implement our own GPU-based SDF reconstruction system in C++ with custom CUDA kernels.
We call this method \textit{KinectFusion+} as the first stage fuses depth images into an occupancy grid (much like KinectFusion \cite{Newcombe:etal:ISMAR2011}), before computing the SDF via the Euclidean distance transform (EDT) of the occupancy and inverse occupancy.
We use the efficient EDT from \citet{Felzenszwalb:Huttenlocher:2004} which allows for parallelisation by decomposing the computation along each axis.
This method is again limited by real-time constraints to a voxel size of 7cm.
More details on KinectFusion+ are provided in the supplementary document.

All experiments are run on an Intel i7-8700K CPU with a GeForce RTX 2080 GPU.

\subsection{Evaluation metrics}
We evaluate the reconstruction quality of the SDF using three different metrics.
All metrics are evaluated at $200k$ points computed by randomly sampling both pixels from the frames and then depths along the back-projected rays.
All methods are evaluated at the same points.
Results for iSDF are averaged over 10 runs with different network initialisations; Voxblox and KinectFusion+ are deterministic.

From a mapping perspective, the first metric is the absolute error between the predicted and ground truth SDF:
\begin{equation}
    \textbf{SDF error: } e_\text{sdf}(\vecx) = \lvert \hat{s}(\vecx) - s(\vecx) \rvert
    ~.
    \label{eqn:sdf_error}
\end{equation}
To mimic the collision cost queries in a real-time planning scenario, the second metric is the error in the predicted collision cost.
We use the cost function from CHOMP \cite{Zucker:etal:IJRR:2013} (with $\epsilon = 2m$) which gives higher importance to regions close to and inside surfaces.
The cost function is similar to the hinge loss but with a quadratic section close to the surface:
\begin{equation}
    c (s) = 
    \begin{cases}
        -s + \frac{1}{2} \epsilon &\mbox{if } s < 0 \\
        \frac{1}{2 \epsilon} (s - \epsilon)^2 &\mbox{if } 0 < s \leq \epsilon \\
        0 &\mbox{otherwise.}
    \end{cases}
\end{equation}
With this, our second metric is:
\begin{equation}
    \textbf{Collision cost error: } e_\text{collision}(\vecx) = \lvert c(\hat{s}(\vecx)) - c(s(\vecx)) \rvert
    ~.
\end{equation}
The accuracy and smoothness of the gradient of the SDF is crucial for optimisation-based planners.
Consequently, our last metric is the cosine distance between the predicted and ground truth gradient vectors:
\begin{equation}
    \textbf{Gradient cosine distance: }
    \small{
    e_\text{grad}(\vecx) = 1 - \frac{\nabla_\vecx \hat{s}(\vecx) \cdot \nabla_\vecx s(\vecx)}{\| \nabla_\vecx \hat{s}(\vecx)\|  \|\nabla_\vecx s(\vecx)\|}
    ~.}
\end{equation}

\begin{figure*}[ht]
    \centering
    \includegraphics[width=\linewidth]{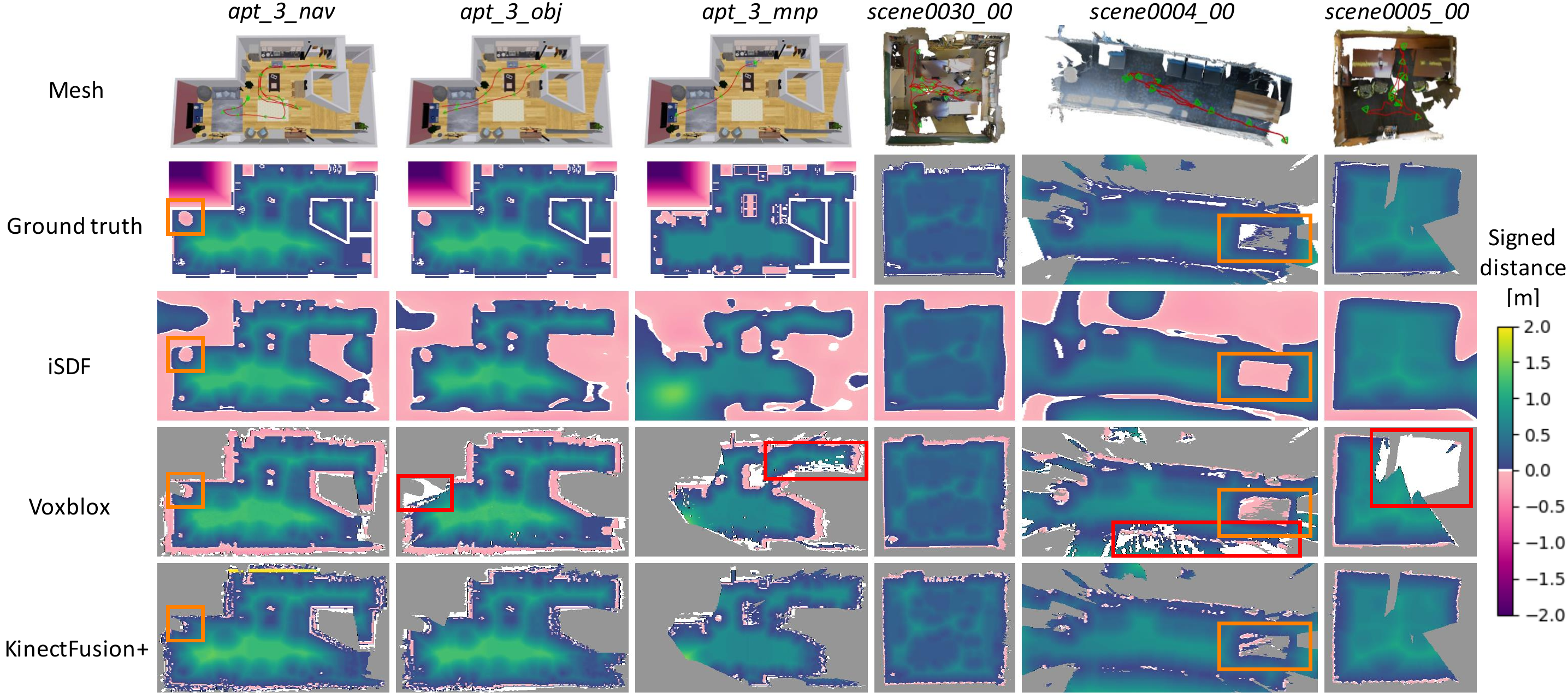}
    \caption{
    \textbf{SDF slices}. Slices at constant height of the reconstructed SDF at the end of the sequence (we choose different heights for each sequence to capture the key scene elements). The meshes are shown for reference with the camera trajectory and keyframes selected by iSDF overlaid. For Voxblox and KinectFusion+, the slices are greyed out in the non-visible region as neither method makes predictions in this region. The ground truth ScanNet \cite{Dai:etal:CVPR2017} slices are greyed out in the non-visible regions as we only have ground truth SDF values in visible regions. White regions in the Voxblox and KinectFusion+ slices are regions that are visible but unmapped (i.e. no rays reach this region).
    }
    \label{fig:main_slices}
    \vspace{-1mm}
\end{figure*}

\vspace{-4mm}
\section{Results}

\begin{figure*}
    \centering
    \includegraphics[width=\linewidth]{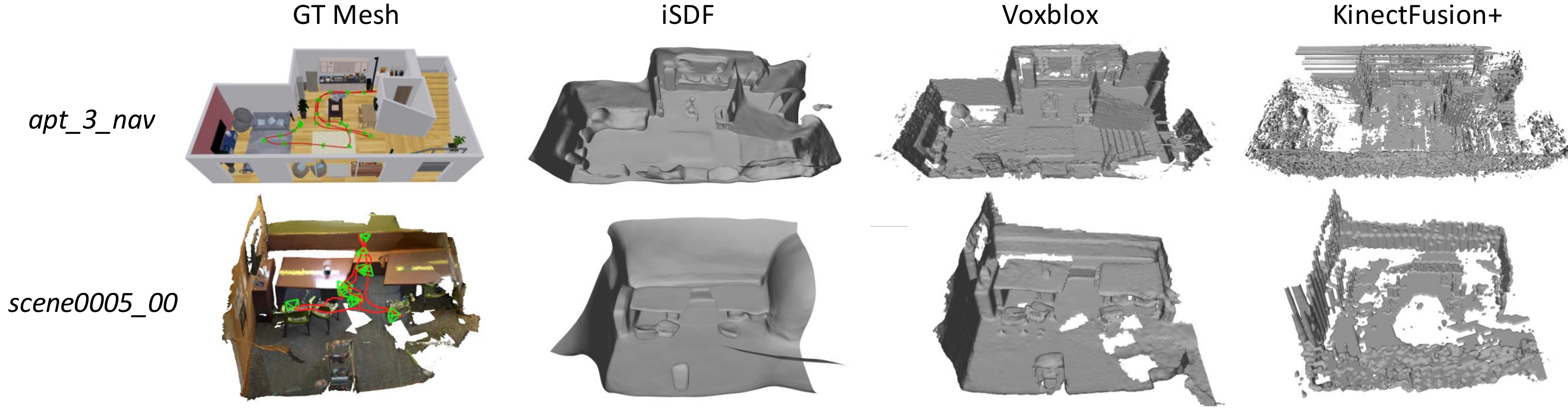}
    \caption{
    The meshes are produced by querying the SDF predictions on a uniform grid and then running marching cubes to find the zero crossing of the level set. iSDF produces a complete and denoised mesh unlike Voxblox and KinectFusion+. 
    }
    \label{fig:level_sets}
    \vspace{-3mm}
\end{figure*}

\begin{figure*}
    \centering
    \includegraphics[width=\linewidth]{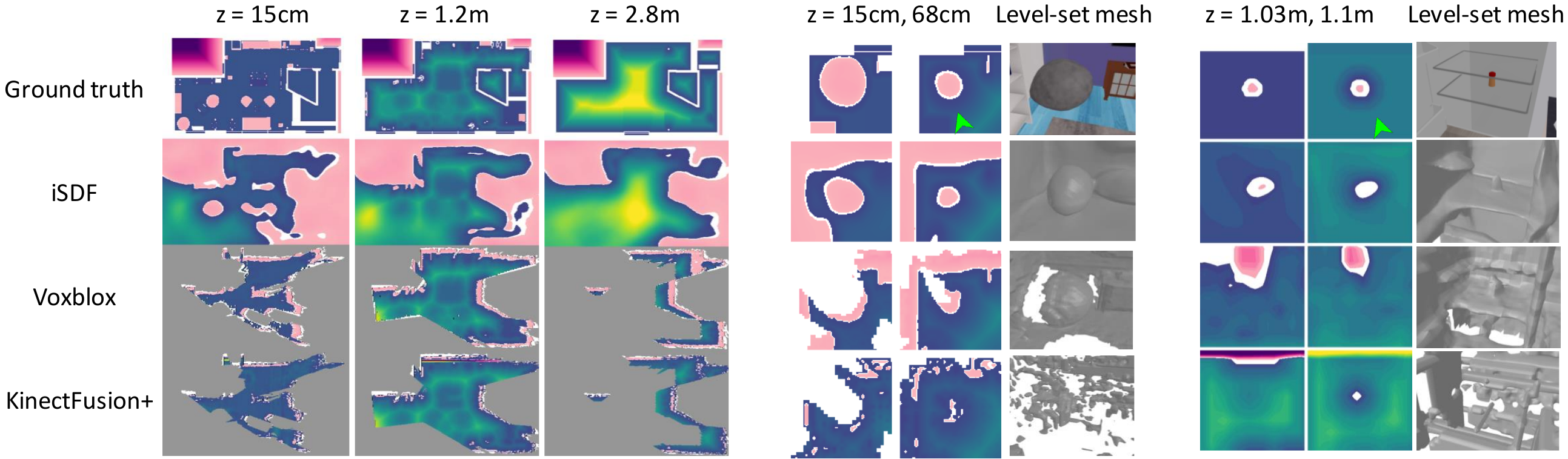}
    \caption{
    All visualisations in this figure are produced using the reconstructed SDFs at the end of a sequence.
    \textit{Left}: The reconstructed SDF at different heights for \textit{apt\_2\_mnp}. The camera is mounted on the robot at a fixed pitch and height of around 1.2m (the height of the middle slices). The other 2 slices are at floor level (15cm) and above the camera (2.8m) and are mostly greyed out in the Voxblox and KinectFusion+ slices as there is limited visibility at these heights.
    \textit{Middle}: SDF slices and the zero level-set mesh for the corner of the room in \textit{apt\_2\_nav} around the beanbag. The mesh is viewed from the location of the green arrow in the ground truth slice.
    \textit{Right}: Slices and zero level-set around the salt shaker for \textit{apt\_2\_mnp}. During the sequence the robot navigates to and approaches the salt shaker, located in the fridge, as if it intends to manipulate it.
    iSDF is the only method that can reconstruct the backside of the beanbag and salt shaker in detail.
    }
    \label{fig:interpolation}
    \vspace{-3mm}
\end{figure*}

We qualitatively and quantitatively evaluate iSDF against Voxblox and KinectFusion+ and find that our method produces more accurate and complete SDFs.
Additionally, due to the choice of a neural field representation, we demonstrate that iSDF can plausibly fill in partially observed regions and map at different levels of detail.
We encourage the reader to view our supplementary video for visualisations of real-time SDF reconstruction.

Throughout the main paper, results are presented for 6 selected sequences (3 synthetic, 3 real) that are representative of all 12 (6 synthetic, 6 real).
Results for the remaining 6 sequences, which provide further evidence for our claims, can be found in the supplementary document.
We evaluate iSDF with the batch distance to the closest point as the bound for the loss unless otherwise stated.

\subsection{Qualitative results}

To visualise the reconstructed SDFs, we compare 2D slices of the SDFs at constant heights in Fig.~\ref{fig:main_slices}.
The first obvious difference is that, unlike Voxblox and KinectFusion+ which map only the visible region, iSDF is predictive and reconstructs a plausible field in the full environment. 
For example, in the interiors of the beanbag and table (highlighted by the orange boxes in Fig.~\ref{fig:main_slices}) or the interior of a wall, iSDF is the only method that returns predictions.

In the main visible regions, all methods produce visually realistic looking fields that are comparable to the ground truth.
There are significant differences however towards the edges of the visible region near surfaces, where iSDF is the only method able to produce a complete reconstruction.
As Voxblox and KinectFusion+ employ fusion in the first stage, there are holes in the reconstructions in regions that no rays have reached, shown in white in the slices. 
This is particularly significant for Voxblox, which to achieve real-time performance on a CPU, discards rays that pass through already updated voxels.
The holes, highlighted by the red boxes in Fig.~\ref{fig:main_slices}, are therefore most common in distant regions that are partially occluded by nearby objects.
These holes are an issue for downstream planners as unmapped regions must be marked as unnavigable and given a high cost.
Additionally, as the gradient field is computed using finite differences, the holes in the gradient field are even larger. 
%

Further evidence that iSDF produces more complete and plausible reconstructions can be seen by visualising the zero level set of the field in Fig.~\ref{fig:level_sets}.
Voxblox and KinectFusion+ produce incomplete and noisy level sets, with bumpy walls and large gaps in the floor.
On the other hand, iSDF is able to sensibly fill in gaps and denoise noisy measurements to produce a smooth and water-tight mesh.
This ability to denoise and interpolate remarkably comes only from priors in the network structure and features learned in a self-supervised manner by fitting the observed part of the room, and without any pretraining.

We further explore the ability of iSDF to plausibly interpolate in partially observed regions in Fig.~\ref{fig:interpolation}.
In the left of the figure, we consider a scenario common for mobile robots in which a camera is mounted at a fixed height and has limited visibility of the floor.
In this case, iSDF is the only method that can make accurate predictions at floor level and above the camera.
In Fig.~\ref{fig:interpolation} \textit{middle}, we demonstrate that due to the global neural representation, in which features are shared for all predictions, iSDF can accurately fill in the unobserved backside of objects.

Unlike uniform voxel grids which set a single resolution for the whole scene, neural fields can adaptively allocate memory capacity to model different parts of the environment with different levels of detail.
This property is very appealing in mobile manipulation settings that require coarse room level reconstructions to navigate towards a target and then fine-detailed reconstructions to manipulate the target.
When reconstructing large environments from a distance, iSDF may miss surface detail that uniform voxel grids can capture, however when the camera approaches an object of interest, increased supervision causes iSDF to allocate more memory capacity to this region and produce fine-detailed object reconstructions.
Fig.~\ref{fig:interpolation} \textit{right} shows results for a sequence in which the robot approaches a salt shaker with the intent to manipulate it. iSDF is the only method capable of reconstructing the small salt shaker while the alternatives based on voxel grids produce coarse and noisy reconstructions.

\subsection{Quantitative results}

\begin{figure*}[ht]
    \centering
    \includegraphics[width=\linewidth]{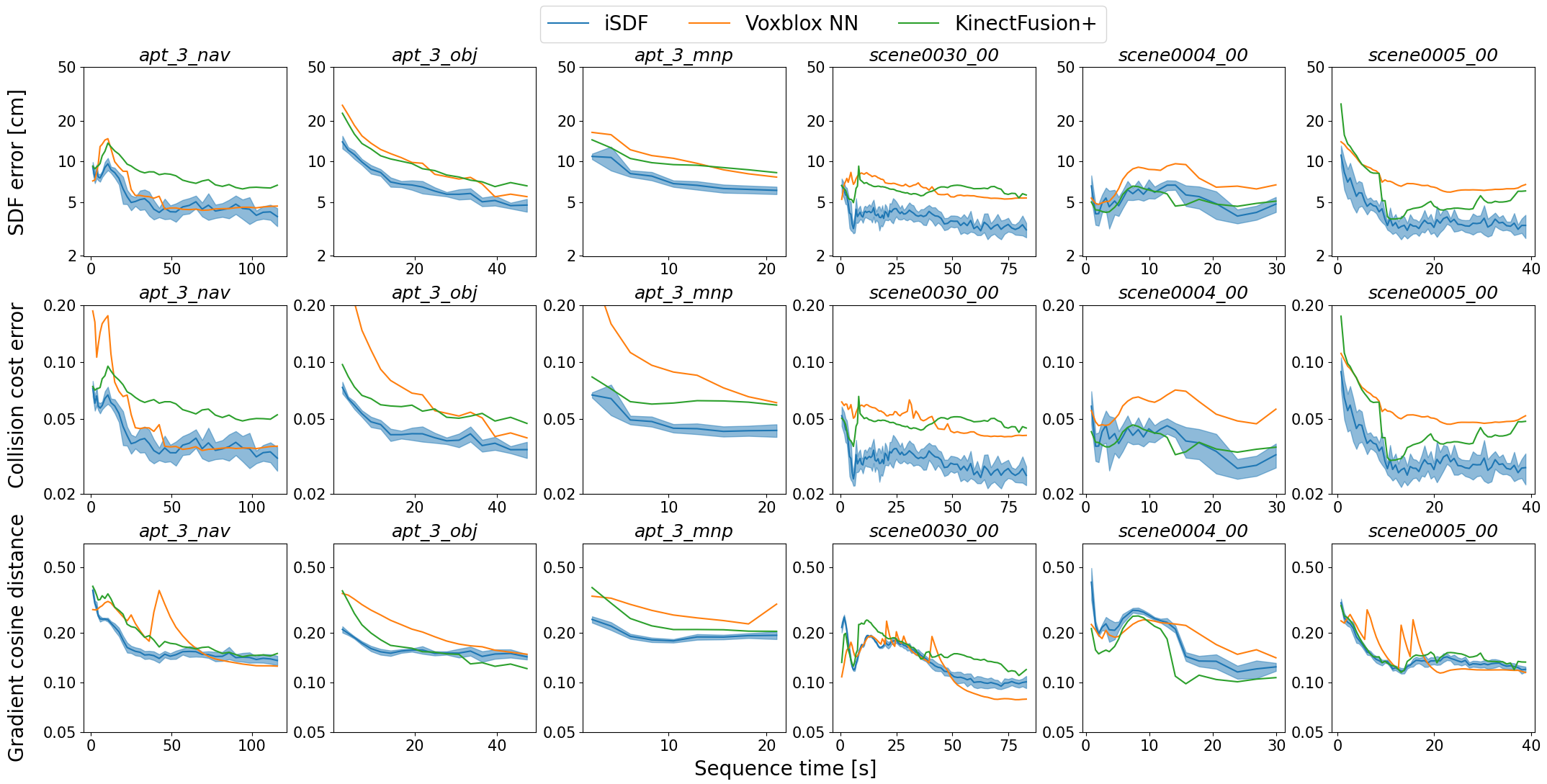}
    \caption{
    We compare iSDF, Voxblox and KinectFusion+ along our three evaluation metrics: SDF error, collision cost error and gradient cosine distance. The metrics are evaluated at regular fixed intervals during the sequences with points sampled in the visible region at that time in the sequence. As Voxblox does not map the full visible region (see Fig.~\ref{fig:main_slices}), we use nearest neighbour interpolation to evaluate the SDF error in unmapped regions. For the collision cost error, we allocate the surface cost $c(0)$ to unmapped regions as a robot would want to avoid unknown regions. In the supplementary document, we present the same plots but with all evaluation points in the Voxblox mapped region and show that there remains a similarly large performance gap between iSDF and other methods.}
    \label{fig:main_seqs}
    \vspace{-2mm}
\end{figure*}

In Fig.~\ref{fig:main_seqs} we present quantitative evaluations along our three evaluation metrics.
iSDF consistently achieves the lowest SDF error, always reaching an error less than $6cm$ at the end of the sequence.
The performance gap varies for different sequences, however iSDF often outperforms the other methods by more than $5cm$.
The collision cost error follows a similar pattern, although notably, the performance of Voxblox is slightly worse along this metric as iSDF and KinectFusion+ are comparatively more accurate close to the surface.
As iSDF produces the most accurate SDF, it is unsurprising that the gradient field is also in general the most accurate.
The performance gap for gradients is however smaller with KinectFusion+ doing better on some sequences.
To understand why iSDF is more accurate, in the supplementary, we split the SDF error into bins according to the true signed distance. We find that far away from surfaces iSDF is significantly more accurate than alternative methods, while very close to surfaces Voxblox can achieve similar accuracy to iSDF. 

The difference in accuracy is primarily due to the large voxel size used by Voxblox and KinectFusion+ for real-time performance.
Voxblox also introduces further errors by measuring distances only along the horizontal, vertical, and diagonal lines in the grid.

A notable result is that iSDF is often better along all metrics by the largest margin at the start of the sequence.
This initial reconstruction is particularly crucial for navigation, as trajectories are more likely to be very poor early on when mapping is limited.
As can be seen in the evolution of the reconstruction in Fig.~\ref{fig:teaser}, iSDF very rapidly captures the rough structure of the room after just a few seconds.

\subsection{Timings and memory}

In real robotics scenarios, memory and compute are often severely limited.
Comparing memory footprints, iSDF requires only 1MB of memory for the network weights and around 20 keyframes to reconstruct a room, while Voxblox and KinectFusion+ typically require 5MB and 30MB respectively to store the voxel grids.
The average time for a single iteration in iSDF is 33ms, with around 2ms for sampling, 2ms for computing bounds, 10ms for the forward pass, 19ms for the backwards pass.

On many robot platforms, compute must be shared between perception, planning, and various other modules.
To study the effects of available compute we also evaluate all methods with half of the compute budget available in the previous real-time experiments.
For iSDF, less compute means fewer optimisation steps per second, while for Voxblox and KinectFusion+ we must increase the voxel size to maintain real-time performance.
We find that iSDF maintains the best performance with half the compute budget, and encourage the reader to see the supplementary document for more details.

\subsection{SDF supervision bounds}
\label{subsec:bounds_ablation}

We validate our choice of using the point batch distance to compute the SDF supervision bounds with an ablation study in the supplementary document.
As expected, we find that tighter bounds lead to more accurate reconstructions.
Using the batch distance gives the lowest SDF error, while the normal correction leads to better accuracy than the ray distance.
There is a trade off between the tightness of the bound and computation time, however, computing the batch distance takes on average 2ms which is minimal in comparison to the 33ms for the full step.

\section{Limitations}
%
%
There remain many challenges in training neural fields more effectively in real-time.
Four important directions are introducing pretrained priors, handling dynamic environments, using local models to reduce replay, and designing more flexible positional embeddings to improve fine-detailed object reconstructions.
This last point is particularly crucial for iSDF as the surfaces can be oversmoothed due to the choice of a low frequency positional embedding ($L=5$) to fit the predominantly low frequency SDF in free space.

Additionally, the Eikonal equation is currently enforced by regularisation which does not handle singularities and leads to over-smoothing around singularities.
An interesting direction would be to investigate constraining the function space to satisfy the Eikonal equation, while accounting for singularities.
Lastly, for a downstream planner, it would be useful to have a measure of confidence associated with the iSDF predictions in unobserved regions.

\section{Conclusion} 

We have presented iSDF, a continual learning system for real-time signed distance field reconstruction.
Key to our approach is a novel self-supervised loss that bounds the predicted signed distance using the distance to the closest batch surface point to enable learning signed distances away from surfaces. 
Positive characteristics of iSDF include adaptive levels of detail, plausible filling in of partially observed regions, denoising of observations, and memory efficient representation.
In evaluations against alternative methods on real and synthetic datasets, we show that iSDF produces more accurate reconstructions as well as better collision costs and gradients for downstream planners.
In the near term, we hope that the generality and differentiability of iSDF means it can easily be integrated into downstream robotics applications from navigation to manipulation.

\section*{Acknowledgments}
We would like to thank Andrew Davison and Zoe Landgraf for insightful discussions as well as Tom Whelan for initial help with datasets. This work was carried out while Joseph Ortiz interned at Meta AI.

\bibliographystyle{plainnat}
\bibliography{robotvision}

\newpage
\section*{\textbf{\large Appendix}}

\section*{KinectFusion+}

Here we describe in detail our baseline KinectFusion+, a GPU based method for real-time SDF reconstruction. KinectFusion+ operates in two stages. In the first stage, we use our own implementation of KinectFusion~\cite{Newcombe:etal:ISMAR2011} to fuse depth measurements into an occupancy grid.

In the second stage, the occupancy grid is transformed to a signed distance field using the Euclidean Distance transform (EDT). The signed distance field is computed by subtracting the EDT of the inverse occupancy field from the EDT of the occupancy field. We employ the efficient EDT algorithm from \citet{Felzenszwalb:Huttenlocher:2004} which has time complexity $O(n)$, where $n$ is the grid size. This algorithm first computes the squared EDT before taking the square root using the insight that the computation of the squared EDT can be separated out along each dimension.

One important detail in KinectFusion+ is the choice to initialise all voxels as unoccupied. With this choice, regions inside objects that cannot be directly viewed will remain as unoccupied and SDF values in these regions should not be used in downstream tasks. In this way, much like Voxblox, KinectFusion+ only maps the shells of objects and cannot be queried inside objects. The alternative of initialising voxels as occupied is overly conservative and leads to highly inaccurate SDF values, although the interiors of objects can now be queried. The reason for the inaccurate predictions is that in free space, the closest occupied voxel may now be an unobserved voxel, meaning the signed distances are greatly underestimated. 

\section*{Additional Results}

\subsection{Reduced Compute Budget Experiment}

\begin{figure}
    \centering
    \includegraphics[width=\linewidth]{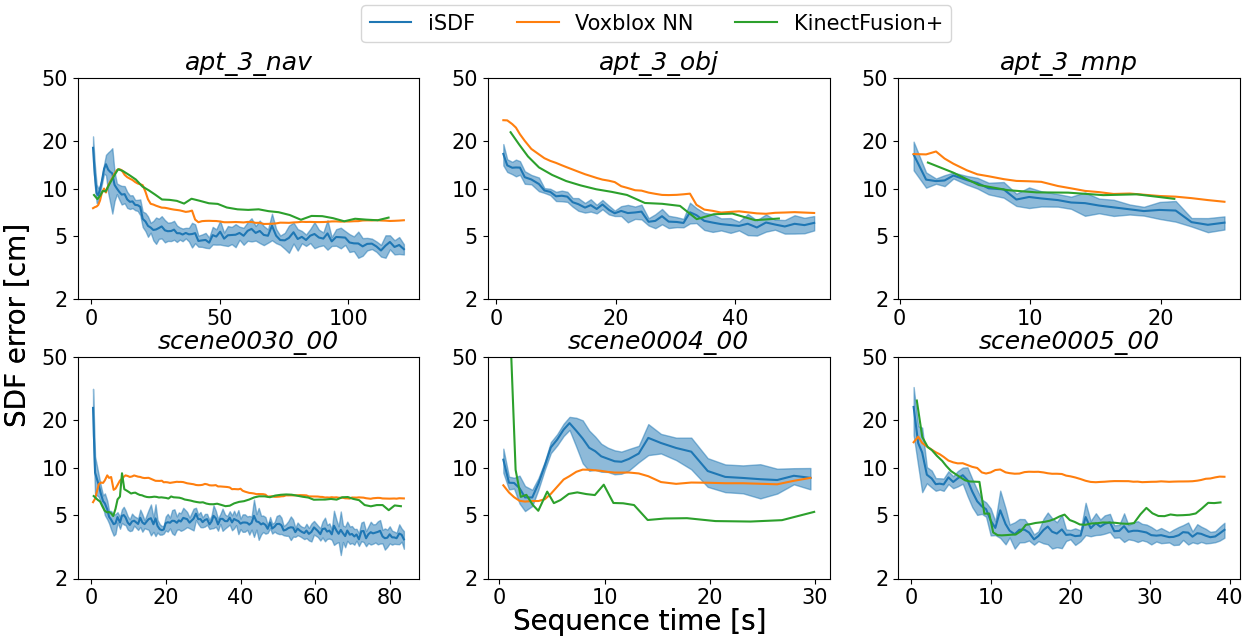}
    \caption{ We compare the SDF error for iSDF, Voxblox and KinectFusion+ when the compute budget is halved.}
    \label{fig:perception_time}
\end{figure}

On many robot platforms, the compute budget must be shared between perception, planning and other modules.
To simulate this constraint, in Fig.~\ref{fig:perception_time} we compare the SDF error when the compute budget available in the main real-time experiments is halved. With half the compute budget, the training speed of iSDF is reduced from around 30 iterations per second to 15. With a reduced compute budget, Voxblox and KinectFusion+ must operate with larger voxel sizes to maintain real-time performance. For Voxblox the voxel size increases from 5.5cm to 7.8cm, and for KinectFusion+ it increases from 7cm to 8cm.

In this reduced compute domain, we find that iSDF retains the best performance on 5 of the 6 sequences. For \textit{scene0004\_00}, which observes a particularly large environment in a short time, the iSDF reconstruction is less accurate.

\subsection{SDF Supervision Bounds ablation}

\begin{figure}[ht]
    \centering
    \includegraphics[width=\linewidth]{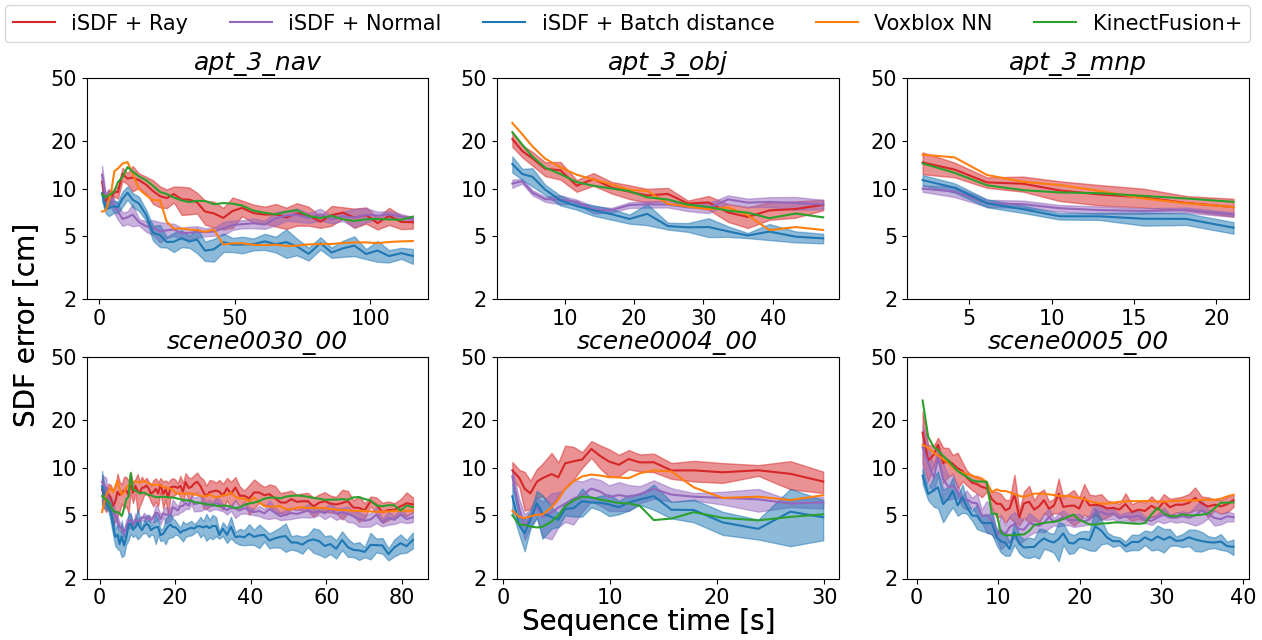}
    \caption{SDF error for iSDF with the different supervision bounds and the baseline methods. 
    As expected, performance improves with tighter bounds from ray, to normal to batch distance.
    iSDF with the point batch distance for the bound gives the lowest error.}
    \label{fig:supervision_ablt}
    \vspace{-2mm}
\end{figure}

To validate our choice of using the point batch distance to compute the SDF supervision bounds we conduct an ablation study in Fig.~\ref{fig:supervision_ablt}.
We find that the point batch distance produces the most accurate SDF reconstruction for all sequences.
Using the normal correction improves on the ray bounds but to a lesser extent than using the batch distance.
These results are expected as the normal correction generally gives tighter bounds than the ray distance while the batch distance bounds are guaranteed to be at least as tight as the ray bounds.

There is a trade off between the tightness of the bound and computation time.
The average time taken to compute the supervision bounds for the ray, normal and batch distance methods is: 0.2ms, 0.6ms, 2.3ms. This time is negligible in comparison to the average total iteration time of 33ms, meaning that using the batch distance bounds has little effect on the number of steps per second.

\subsection{Evaluation in Voxblox mapped region}

As discussed in the main paper, the Voxblox reconstruction has significant holes in the visible region. These holes are caused by discarding rays during TSDF fusion that pass through voxels that have already been updated by rays in the same image. This simplification, which is necessary for real-time performance on a CPU, causes large holes in regions distant from the camera that are partially occluded by foreground objects.

\begin{figure*}
    \centering
    \includegraphics[width=\linewidth]{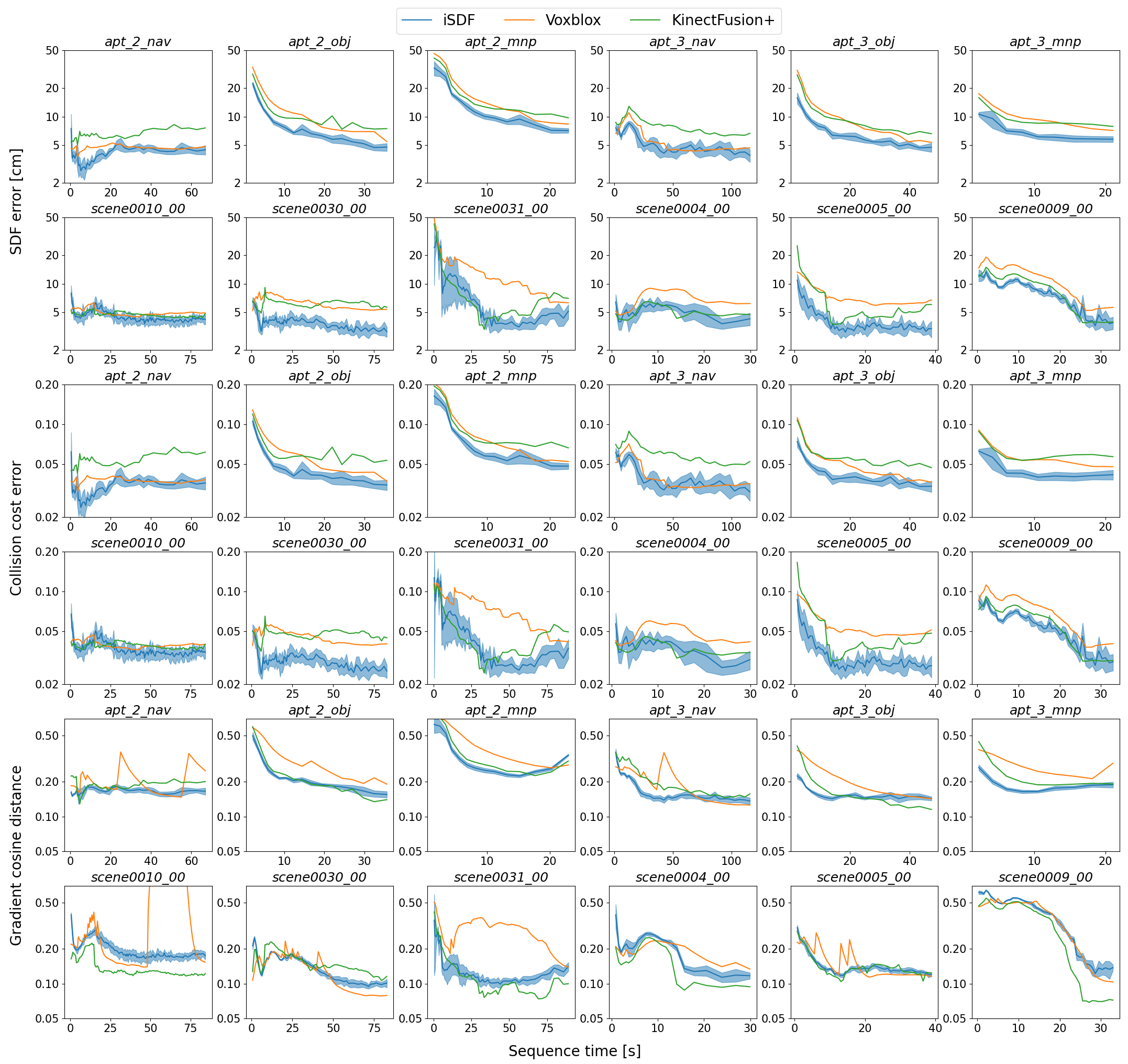}
    \caption{
    We compare the SDF error evaluted only at points in the Voxblox mapped region. As a result, we discard points from the evaluation in the most challenging regions to reconstruct that are heavily occluded and distant from the camera.
    }
    \label{fig:vox}
\end{figure*}

One way to avoid these holes would be to use a GPU based TSDF fusion method in the first stage, that does not discard rays, and then use Voxblox's wavefront propagation algorithm to transform the resulting TSDF to SDF. To approximately evaluate against such a hypothetical system, we compare the existing methods along our three metrics evaluated at points only in the Voxblox mapped region in Fig. \ref{fig:vox}. The evaluation is conducted in the same manner as in the main paper by sampling along rays in the visible region, however points outside of the Voxblox mapped region are discarded.

In the Voxblox mapped region, iSDF still produces the most accurate SDFs and best approximations of collision costs. 
As expected, the performance of Voxblox improves relative to the other methods, however it still records higher SDF error than iSDF on all sequences. 
Voxblox also produces the least accurate gradients, with iSDF and KinectFusion+ producing similarly accurate gradient fields in this region.

\subsection{SDF error binned by distance to the surface}

\begin{figure*}
    \centering
    \includegraphics[width=\linewidth]{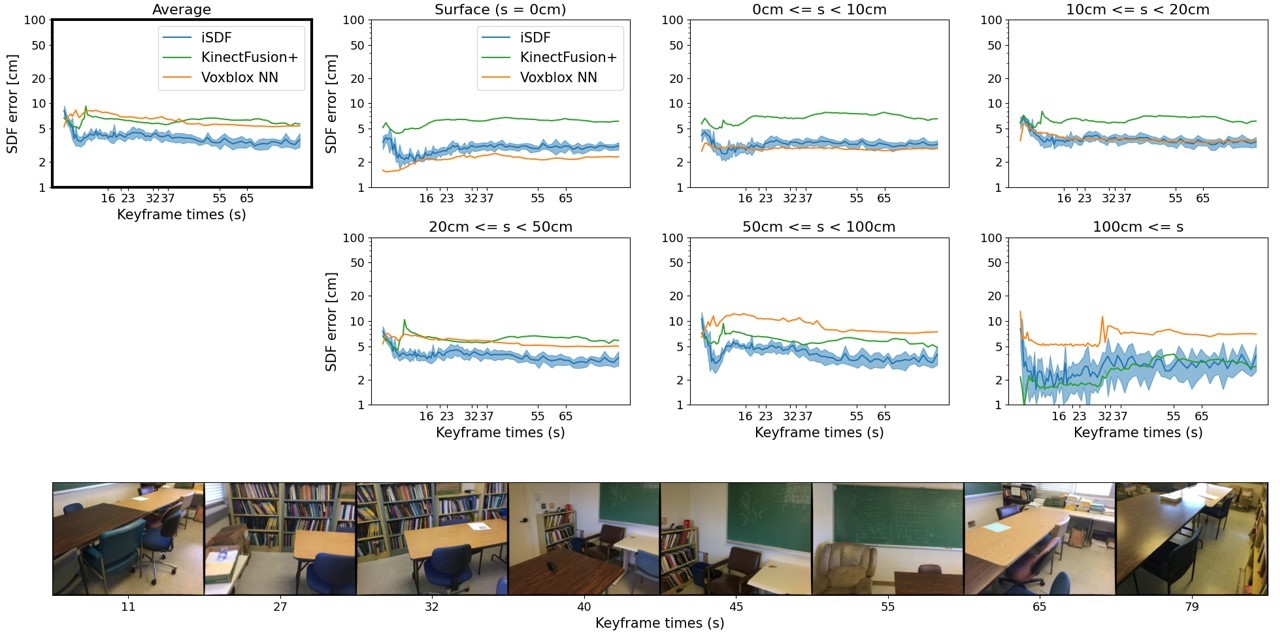}
    \caption{
    We break down the average SDF error into bins according to the true signed distance ($s$). There are no bins for negative SDF values as for the ScanNet dataset we can only compute ground truth signed distances in regions outside of objects. In the bottom row, we show 8 of the keyframes selected by iSDF during the sequence \textit{scene0030\_00}. 
    }
    \label{fig:bins}
\end{figure*}

To understand why iSDF achieves a lower SDF error than Voxblox and KinectFusion+, we break down the SDF error into bins according to the true signed distance for \textit{scene0030\_00} in Fig \ref{fig:bins}. We observe that far from surfaces iSDF and KinectFusion+ are significantly more accurate than Voxblox, while close to surfaces iSDF and Voxblox are the most accurate methods. iSDF therefore has the lowest average SDF error as is the only method that has low SDF error both close to and far from surfaces. Although we only show results here for one sequence, other sequences show a similar trend.

\subsection{Results for 6 additional sequences}

In this section, we present results for the 6 sequences (3 ReplicaCAD~\cite{Szot:etal:ARXIV2021} and 3 ScanNet~\cite{Dai:etal:CVPR2017} sequences) that were not used in the main paper. These figures provide further evidence for the claims made in the paper and so we simply show them here and refer the reader back to the main paper for the discussion.

\begin{figure*}
    \centering
    \includegraphics[width=\linewidth]{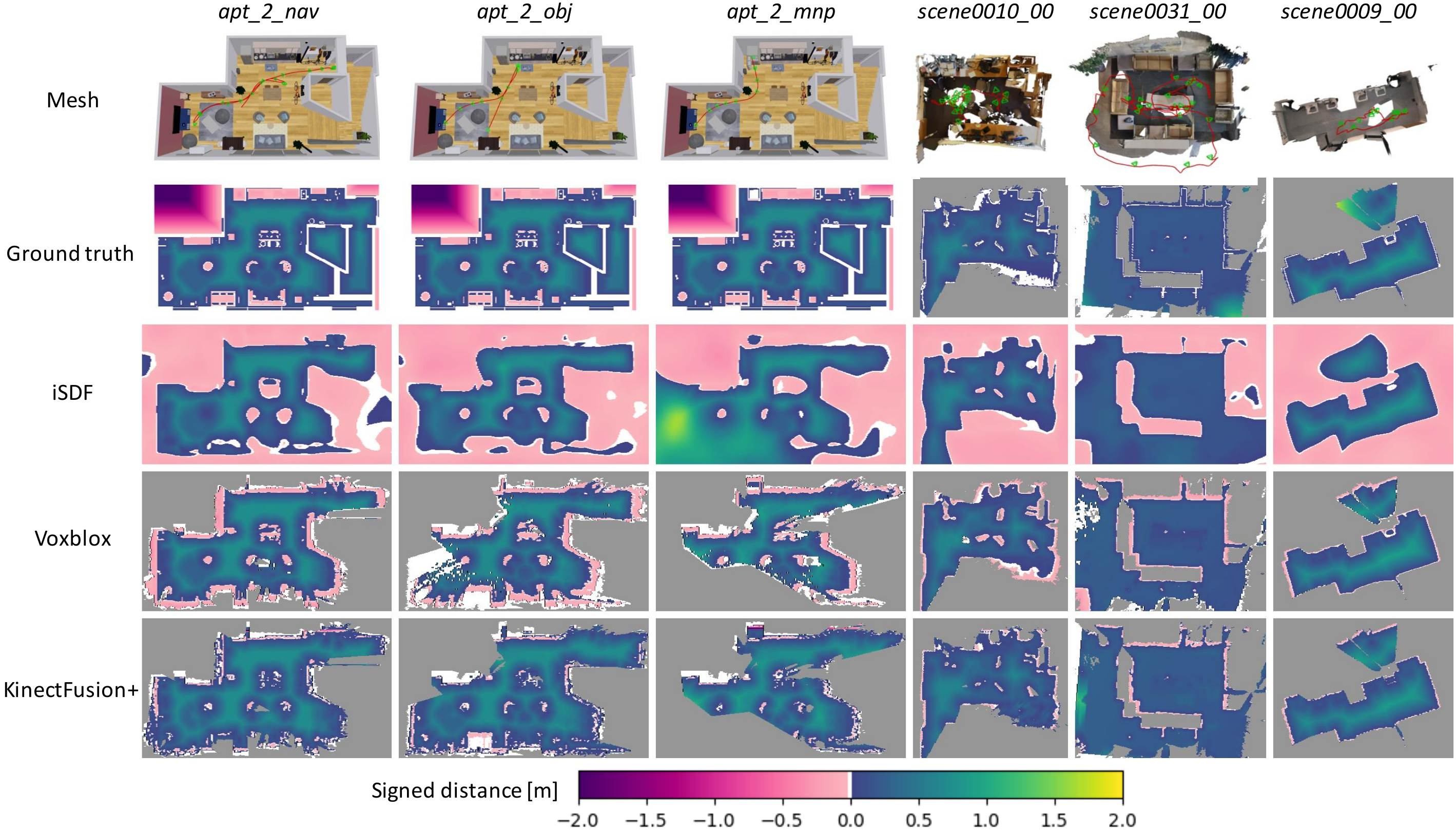}
    \caption{
    \textbf{SDF slices}. Slices at constant height of the reconstructed SDF at the end of the sequence. 
    The meshes are shown for reference with the camera trajectory and keyframes selected by iSDF overlaid. For Voxblox and KinectFusion+, the slices are greyed out in the non-visible region as neither method makes predictions in this region. The ground truth ScanNet \cite{Dai:etal:CVPR2017} slices are also greyed out in the non-visible region as we only have ground truth SDF values in the visible region. White regions in the Voxblox and KinectFusion+ slices are regions that are visible but that are unmapped (i.e. no rays have reached this region).
    Note that in \textit{scene0009\_00} there is a mirror, causing all methods to reconstruct the reflection of the room.
    }
    \label{fig:slices_supp}
\end{figure*}

\begin{figure*}
    \centering
    \includegraphics[width=\linewidth]{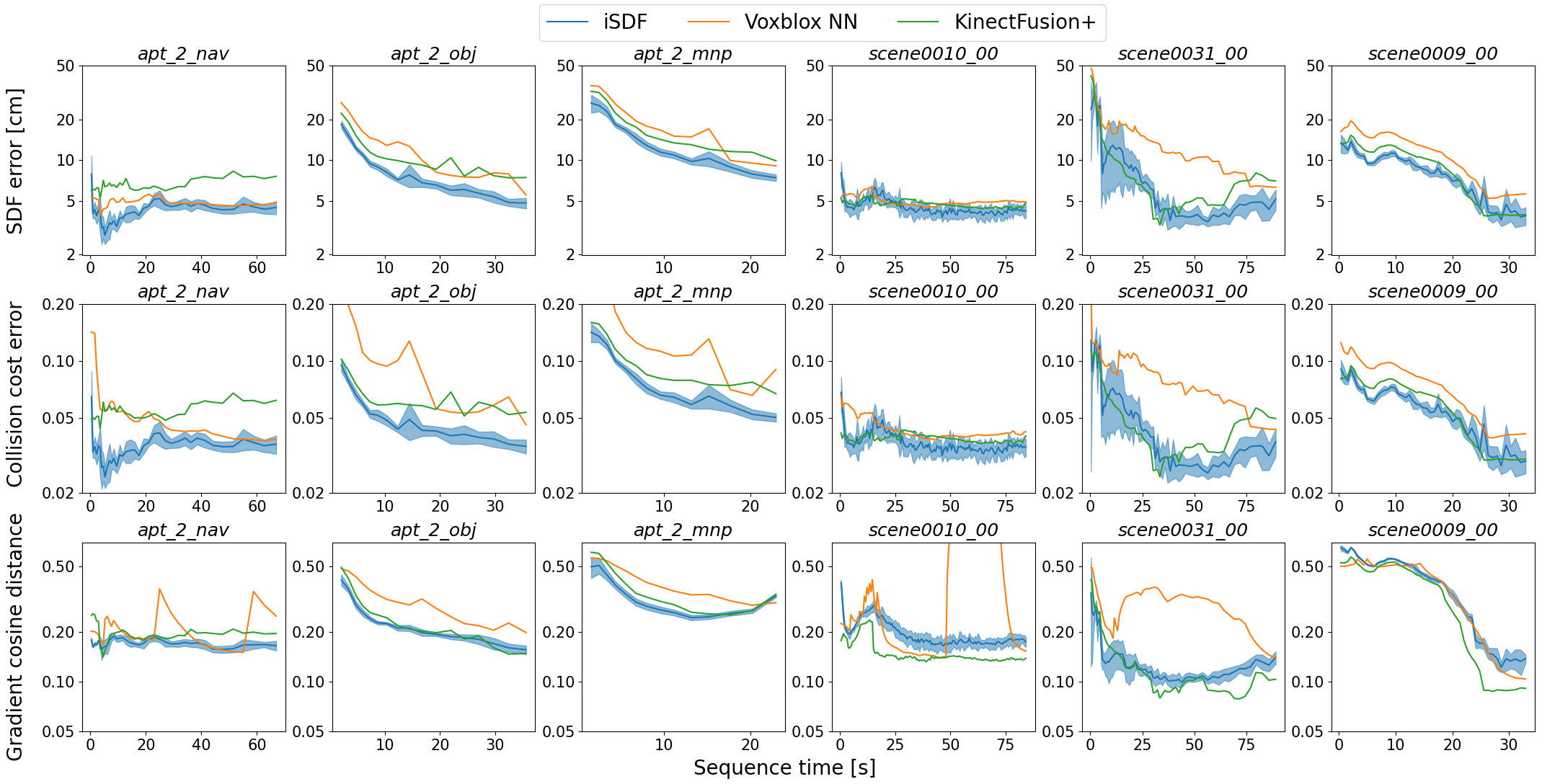}
    \caption{
    We compare iSDF, Voxblox and KinectFusion+ along our three evaluation metrics - SDF error, collision cost error and gradient cosine distance. The metrics are evaluated at regular fixed intervals during the sequences with points sampled in the visible region at that point in the sequence. As Voxblox does not map the full visible region, we use nearest neighbour interpolation to evaluate the SDF error in unmapped regions. For the collision cost error, we allocate the surface cost $c(0)$ to unmapped regions as a robot would want to avoid unknown regions. In \textit{scene0010\_00}, Voxblox records very high gradient error in a duration of the sequence when there are many gaps in the Voxblox reconstruction.}
    \label{fig:supp_seqs}
\end{figure*}

In Fig.~\ref{fig:slices_supp} we compare slices at constant height of the reconstructed SDF at the end of the sequence. As in the main paper, we see that iSDF produces more complete and accurate slices.
In Fig. \ref{fig:supp_seqs} we compare the performance along our three evaluation metrics. iSDF consistently has the lowest SDF and collision cost error, while KinectFusion+ produces similarly accurate gradients.

\end{document}